# Put your money where your mouth is: Using deep learning to identify consumer tribes from word usage

Gloor, P., Fronzetti Colladon, A., de Oliveira, J. M., & Rovelli, P.

# Put your money where your mouth is: Using deep learning to identify consumer tribes from word usage

Gloor, P., Fronzetti Colladon, A., de Oliveira, J. M., & Rovelli, P.

**Abstract**

Internet and social media offer firms novel ways of managing their marketing strategy and gain competitive advantage. The groups of users expressing themselves on the Internet about a particular topic, product, or brand are frequently called a *virtual tribe* or *E-tribe*. However, there are no automatic tools for identifying and studying the characteristics of these virtual tribes. Towards this aim, this paper presents *Tribefinder*, a system to reveal Twitter users' tribal affiliations, by analyzing their tweets and language use. To show the potential of this instrument, we provide an example considering three specific tribal macro-categories: alternative realities, lifestyle, and recreation. In addition, we discuss the different characteristics of each identified tribe, in terms of use of language and social interaction metrics. *Tribefinder* illustrates the importance of adopting a new lens for studying virtual tribes, which is crucial for firms to properly design their marketing strategy, and for scholars to extend prior marketing research.

# 1. Introduction

A tribe is "a network of heterogeneous persons linked by a shared passion or emotion" (Cova & Cova, 2002, p. 602). Put differently, it is a means whereby individuals experience a sense of community and share strong emotional links, common culture, passions, and vision of life (Cova, 1996; Cova & Cova, 2002; Richardson, 2013). Individuals break up in several mainstream tribes; at the same time, each individual may belong to many smaller and larger tribes, playing different roles and wearing different masks (Cova, 1996; Cova & Cova, 2002).

The emergence of the concept of *tribalism* in management science (Bauman, 1990; Maffesoli, 1996) made it clear that *tribes* are fundamental for firms' survival (e.g., Holzweber, Mattsson, & Standing, 2015), being especially important for marketing (e.g., Goulding, Shankar, & Canniford, 2013; Kozinets, 1999). Their characteristics (e.g., preferences, aptitudes, behaviors, needs) strongly affect the success of a marketing campaign (Cova & Cova, 2002). There is widespread agreement among marketing scholars that firms should extend traditional marketing strategies (Addis & Podesta, 2005; Canniford, 2011) by designing marketing actions based on the existence and behavior of the consumer tribes that the firm has to address given its characteristics, its brand, and the product or services it offers (Moutinho, Dionísio, & Leal, 2007). Understanding which tribes are particularly attracted by a specific brand, product or service, and then studying their characteristics, is a powerful instrument to improve marketing campaigns. In doing so, firms may design their marketing actions according to the individual and social needs of tribes' members (e.g., Cova, 1996; Holzweber, et al., 2015), thus maximizing their probability of success. This approach is known as *tribal marketing* (Cova & Cova, 2002).

Despite its potential benefits, to the best of our knowledge only a few studies have investigated how tribes can be effectively used as a strategic resource (Cova & Cova, 2002).

Most difficulties are encountered by researchers and business analysts in the process of data collection and tribe classification. Specifically, an instrument apt to automatically and systematically identify tribes in the current digital era is still missing. Nowadays, the identification of tribes requires different and special efforts (Cova & Cova, 2002), as the advent of Internet and the growing use of social media as marketing instruments pose new challenges (Burton & Soboleva, 2011; Dwivedi, Kapoor, & Chen, 2015). Social media platforms are more and more used as marketing instruments (e.g., Burton & Soboleva, 2011; Cova & White, 2010; Dwivedi, et al., 2015) as they allow companies to reach a wider group of (potential) customers (Weinberg, 2009). This is confirmed by the raise of *social media marketing*, and by the fact that social media often became more prominent than other advertising means (e.g., Dwivedi, et al., 2015). This shift, from *offline* to *online* competition, challenges both the definition and identification of tribes. We thus witness a change in the concept of tribes, which are now more properly referred to as *virtual tribes* or *E-tribes*, meaning tribes that nowadays form by communication technologies (Cova & Pace, 2006; Hamilton & Hewer, 2010). This calls for new methodologies to properly identify and study these virtual tribes.

Our paper aims to address this methodological gap, by introducing a novel system leveraging latest developments in AI, called *Tribefinder*, developed to identify virtual tribes. Analyzing information published by users on social media, *Tribefinder* is able to systematically categorize them in alternative tribes. This is possible as individuals' expressed behaviors reveal to what tribes they belong and how they perceive their own identity (e.g., Garry, Broderick, & Lahiffe, 2008). Indeed, each tribe has its own peculiarities, behaviors, rituals, traditions, myths, values, beliefs, hierarchy, and vocabulary (Cova & Pace, 2006), which support the identification of individuals' tribal affiliations. In its current version, *Tribefinder* relies on textual data to automatically assess individuals' tribal affiliations through the analysis of their generated content. Among the freely accessible text data sources, Twitter is one of the most popular. The user generated content extracted from this platform offers a direct window into consumers' feelings and opinions on products (Burton & Soboleva, 2011; Lipizzi, Iandoli, & Marquez, 2015; Martínez-Rojas, del Carmen Pardo-Ferreira, & Rubio-Romero, 2018), which are fundamental for firms to infer their attitude towards brands (Mostafa, 2013; Shirdastian, Laroche, & Richard, 2017). Indeed, Twitter has become one of the most important places where firms develop and test their marketing strategies (e.g., Burton & Soboleva, 2011; Jansen, Zhang, Sobel, & Chowdury, 2009; Singh, et al., 2017b; Thomases, 2009). Accordingly, the *Tribefinder* has been developed to work with Twitter data. Nevertheless, the system could be easily configured to operate with other text sources, such as email communications or data extracted from other social media platforms. *Tribefinder* works as a combination of word embeddings and LSTM machine learning models (Gloor, et al., 2019) and is flexible in the definition of many possible tribal categories. In this paper, we describe the basic functions of *Tribefinder* and provide some examples of its use. We show how *Tribefinder* is currently able to classify Twitter users according to three tribal macro-categories: *alternative realities*, *lifestyle*, and *recreation*. We adopted this tribal categorization to be in line with an earlier work that used the *Tribefinder* (Gloor, et al., 2019). While these categories can capture some important characteristics of the individuals that typically interact on social media, it is worth mentioning that the system can be easily reconfigured to work with other tribal macro-categories. After describing the *Tribefinder* functionality, we present a study of the characteristics of tribes resulting from tweets related to the discourse about major international brands.

This paper advances past and current research by introducing a new instrument to easily identify tribal affiliations. Moreover, it provides useful insights to study the characteristics and the interaction dynamics of different tribes. In this regard, our work mostly contributes to the marketing stream of research. Our instrument may constitute a foundation to advance research on the strategic use of marketing actions (e.g., Dwivedi, et al., 2015; Weinberg, 2009), which is currently quite limited (Cova & Cova, 2002). As it is typically costly to define one-to-one marketing actions to firms' potential customers, *Tribefinder* allows to identify and isolate coherent segments of customers, helping in the customization of advertising campaigns and interactions with tribe members. Accordingly, suggestions may be derived to extend traditional marketing solutions (Addis & Podesta, 2005; Canniford, 2011) and to inform firms on new marketing solutions able to incorporate the essence of the tribes interested in the services they offer (Moutinho, et al., 2007). The development of this new system also adds to current researches on social media marketing (see Dwivedi, et al., 2015 for a review of the literature) adopting modern methodologies to analyze tweets for marketing purposes (e.g., Fronzetti Colladon, 2018; Gloor, 2017; He, Zha, & Li, 2013), which have never been applied to the identification of tribal affiliations. Moreover, our work adds to the general literature on social media mining (e.g., Alalwan, Rana, Dwivedi, & Algharabat, 2017; Grover, Kar, Dwivedi, & Janssen, 2018; Jeong, Yoon, & Lee, 2017; Singh, Dwivedi, Rana, Kumar, & Kapoor, 2017a), offering a new system to easily analyze and extract information about social media users. Finally, we are convinced that *Tribefinder* can be of high value for firms that want to develop a better understanding of their brand's virtual tribes, to measure the efficiency of their marketing campaigns, and to set up or adjust their marketing strategies, thus improving their competitive advantage and performance.

## 2. Theoretical background

### 2.1. The concept of tribe in marketing

Throughout the years, marketing scholars have become more and more interested in understanding, organizing and facilitating brand communities (Algesheimer, Dholakia, & Herrmann, 2005; Casaló, Flavián, & Guinalíu, 2010; McAlexander, Schouten, & Koenig, 2002), which are defined as "a specialized, non-geographically bound community, based on a structured set of social relationships among users of a brand" (Muniz & O'Guinn, 2001, p. 412). Brand communities are important for several reasons; for instance, they are able to influence their members' perceptions, thoughts, and actions (e.g., Muniz & Schau, 2005). At the same time, firms can exploit them to disseminate relevant information rapidly (e.g., Brown, Kozinets, & Sherry Jr, 2003), acquire feedback from consumers on products, new offerings, and competitive actions, and engage and collaborate with their most loyal customers (e.g., Franke & Shah, 2003). Customers may in turn contribute to increase a firm's value and competitive strategy, contributing to firms' innovation processes (Etgar, 2008; Franke & Piller, 2004; Prügl & Schreier, 2006; Von Hippel, 2005); the same may happen for the marketing strategy that may benefit from customer feedback (Lusch & Vargo, 2006). Therefore, understanding brand communities is of pivotal importance for firms.

Brand communities can be considered as complex systems in which members – despite sharing the same interest in a specific brand, product or service – may differentiate from each other in several characteristics, pertaining for instance to their individual traits, aptitudes, general interest, or preferences. While a brand community is tied specifically to a brand and explicitly commercial, we know from literature that its members may belong to one or more *tribe communities* or *tribes* (Goulding, et al., 2013; Ryan, McLoughlin, & Keating, 2006). These are not tied to a specific brand, but rather to a specific set of values, and their members engage in social actions that are facilitated by a variety of brands, products, activities and services (Cova & Cova, 2002).

According to Cova (1996), the postmodern society consists of micro groups of individuals that share strong emotional links, a common sub-culture, and a vision of life, which leads to the creation of invisible tribes. The word *tribe* recalls quasi-archaic values, such as a local sense of identification, grandiosity, syncretism, and group narcissism (Cova, 1997; Cova &

Cova, 2002). More specifically, a tribe is defined as "a network of heterogeneous persons – in terms of age, sex, income, etc. – who are linked by a shared passion or emotion" (Cova & Cova, 2002, p. 602). Through their behaviors, individuals self-select the tribes to align with (Holzweber, et al., 2015). In this respect, self-categorization theory indicates that individuals classify themselves into social categories in order to favor a self-definition in the social environment to which they belong (Haslam & Turner, 1992). Among the motivations that push individuals to self-categorization there is the need of increasing self-esteem, social identity and social prestige (Ellemers, Kortekaas, & Ouwerkerk, 1999). Moreover, tribal members develop a sense of community and solidarity, ethnocentrism, devotion, emotional connection, secrecy, and sustenance of the collective (Garry, et al., 2008; Maffesoli, 1996). For these reasons, individuals recognize themselves as part of one tribe or, in the majority of cases, of different tribes (Elliott & Davies, 2006). Indeed, being member of one tribe does not prohibit the identification with other tribes (Goulding, et al., 2013). At the same time, tribe members are recognized from the outside as members of a group (Garry, et al., 2008) and often meet and perform rituals in public places, called "anchoring events" (Aubert-Gamet & Cova, 1999), as well as conveying (visible and invisible) signs to facilitate their identification with the tribes (Veloutsou & Moutinho, 2009). Finally, within tribes individuals may play different roles and wear different masks (Cova, 1996; Cova & Cova, 2002). We are thus witnessing the emergence of *tribalism* (Bauman, 1990; Maffesoli, 1996), which is characterized by a mutual relation existing between each tribe member and the tribe as a collective whole. On the one hand, tribe members define the tribe. On the other hand, the tribe influences tribe members' behaviors.

Following this rather new concept of tribe, scholars suggested the creation of new postmodern marketing techniques. Current research is highlighting the value of tribes in this context (Arnould & Thompson, 2005; Kozinets, 1999). For instance, Holzweber, et al. (2015)

found that understanding tribes' development is critical for the survival of entrepreneurial ventures. Extending their arguments, tribes may be of particular importance for the success of every kind of firm, from emerging early-stage ones to the most established ones. Firms should try to understand individual and social needs of their brand community members (Cova, 1996; Holzweber, et al., 2015), thus uncovering their tribal affiliations and manage marketing actions specifically designed for their characteristics. Therefore, firms are advised to leverage the presence of tribes impassioned by a brand (Cova & Cova, 2002; Solomon, 2003), in this sense efforts are needed to strengthen good marketing practices (Broderick, MacLaran, & Ma, 2003). This approach is known as *tribal marketing* (Cova & Cova, 2002).

Nevertheless, there is still much to learn about how to build and maintain tribes, how to identify consumers' tribal affiliations, and how to exploit tribes for marketing purposes (Goulding, et al., 2013). Few studies made an attempt to explain how tribes may be leveraged as a strategic resource (Cova & Cova, 2002). We contend that this gap at least partially depends on the limited ability of traditional instruments to identify tribes and their characteristics. This has been made even more difficult by the advent of new digital media, which changed the paradigm of simple tribes to *virtual tribes* or *E-tribes* (Cova & Pace, 2006; Hamilton & Hewer, 2010).

### 2.2. From "tribe" to "virtual tribe" via Twitter

The growth of new digital media in the first decade of the twenty-first century contributed to transforming marketing communications (e.g., Burton & Soboleva, 2011; Casaló, et al., 2010; Newman, Chang, Walters, & Wills, 2016). Specifically, the advent of the Internet and of the Web 2.0 technologies is increasingly fueling tribal marketing (Cova & White, 2010). Nowadays, on social media platforms, individuals with common interests gather for sharing, conversation, and interactions (Grover, et al., 2018; Hamilton & Hewer, 2010; Jeong, et al., 2017; Singh, et al., 2017a; Weber, 2007). These social media platforms thus serve as additional communication channels that firms can exploit for management (Martínez-Rojas, et al., 2018) and marketing purposes (Alalwan, et al., 2017; Burton & Soboleva, 2011; Dwivedi, et al., 2015; Jeong, et al., 2017; Kapoor, et al., 2018). Among the others, Twitter plays a key role (e.g., Grover, et al., 2018).

Launched in 2006, Twitter rapidly became one of the most popular social media platforms and the dominant microblogging social networking service (Ameen & Kaya, 2017; Case & King, 2011). Over time, it revealed its importance for firms (Hennig-Thurau, Wiertz, & Feldhaus, 2015), which may use it to reach their current and potential customers, advertise their products, and get relevant information for marketing purposes (Burton & Soboleva, 2011; Jansen, et al., 2009; Singh, et al., 2017b; Thomases, 2009). If properly exploited, Twitter may thus serve as a potent weapon to generate competitive advantage and improve performance (Grover, et al., 2018). On Twitter, individuals openly reveal their feelings and opinions on products (Jeong, et al., 2017; Lipizzi, et al., 2015), and their tweets help detecting their attitude towards brands (Mostafa, 2013; Shirdastian, et al., 2017). The interactions that take place on this social media platform, both among individuals and between firms and their consumers, may induce other individuals to become consumers and affect firms' performance and their development of new products (e.g., Wang & Sengupta, 2016), based on the feedback they receive on the platform.

Given its importance, Twitter is a main contributor to change the paradigm of tribes, leading to the new concept of *virtual tribes* or *E-tribes* (Cova & Pace, 2006; Hamilton & Hewer, 2010). Virtual tribes are of key importance for the success of marketing actions and business strategies (e.g., Moutinho, et al., 2007). Kozinets (1999) suggests that if marketers succeed in understanding in a rigorous way their consumers' virtual tribes, how consumers interact within these tribes, and their characteristics, they will be able to better execute marketing actions and improve their firm's performance. Indeed, marketing actions may generate feelings, thoughts, attitudes and experiences that can influence consumers' response and purchase intentions (Keller, 2016), which may then reflect their social activity on Twitter, thus providing a direct feedback to marketers. This feedback may then be used to adjust marketing actions, as well as to develop new product enhancements and ideas (Kozinets, 1999).

## 3. A new instrument

### 3.1. The challenge of identifying tribes

The study of tribes has typically been done by ethnographic and nethnographic approaches (Cova & White, 2010; Goulding, et al., 2013; Hamilton & Hewer, 2010), focus groups (Dionísio, Leal, & Moutinho, 2008; Moutinho, et al., 2007), interviews (Cova & Cova, 2002; Cova & Pace, 2006; Holzweber, et al., 2015), and surveys (Taute & Sierra, 2014). While these approaches may provide a deep understanding of tribes' dynamics, they do not allow to automatically and systematically identify virtual tribes and their characteristics. The identification of the tribal community thus remains a major challenge; so far, many researchers worked based on convenience samples of individuals, which did not sufficiently represent the population at large of a firm's consumers. As an example, to select the participants of their focus groups, Moutinho, et al. (2007) recruited individuals directly on a surfing beach by a basic filtering questionnaire about their level of involvement with this sport. Similarly, Mitchell and Imrie (2011) limited their case study to a single self-formed group of adult record collectors. In light of the revolution initiated by the widespread use of the Internet and Twitter for marketing purposes, these methodologies are no longer suitable for the new context towards which research and competition are moving, and the related new goals posed by scholars and firms.

### 3.2. The Tribefinder solution

To solve the limits of traditional methodologies, we propose a novel systems that exploits the freely accessible text data made available by Twitter (Bringay, et al., 2011; Fronzetti Colladon, 2018; Grover, et al., 2018; He, et al., 2013) to offer tribal categorizations for the benefit of both researchers and firms. Specifically, *Tribefinder* aims at categorizing Twitter users into different tribal categories – i.e. a user's language might partially match multiple tribes but, within each tribal macro-category, users have one single predominant affiliation. This is done by analyzing tweets to extract information about key individuals, brands, and topics. The output consists of the tribal affiliations of the individual whose tweets are analyzed. In its first applications, the *Tribefinder* was used to classify individuals along three specific tribal macro-categories: alternative realities, lifestyle, and recreation, which allow to capture some individual differences that Twitter's users typically tend to (consciously and unconsciously) communicate on this social media. Nevertheless, these three specific macro-categories serve as examples to explain how *Tribefinder* works. This is not intended to be a limitation. Indeed, *Tribefinder* can be easily customized according to the interests of who uses it and new tribes can be added. Applying the same methodology explained below in detail, *Tribefinder* is currently able to identify arbitrary user-defined tribal macro-categories.

As shown in Figure 1, *Tribefinder* consists of two modules: tribe creation and tribe allocation. To build and then train the *Tribefinder* system, the user has first to identify a sample of key individuals who represent the different tribes that s/he has selected for each tribal macro-category (for instance, in our specific example, the tribes *nerd*, *fatherlander*, *spiritualist*, and *treehugger* for the macro-category *alternative realities*). A large sample of Twitter users, belonging to each of these chosen tribes, is identified. Initial search is based on user-defined keywords, related to concepts, ideas, and artifacts that pertain to each tribe. For instance, for a tribe of "gun-lovers", a few hundred Twitter profiles of self-professed gun aficionados would be selected. As is known from literature, a tribe can be idealized as a concept, idea, or artifact that its members believe in or like (De Oliveira & Gloor, 2018). The *Tribecreator* module (De Oliveira & Gloor, 2018) is used to perform this search. *Tribecreator* is a Web tool that automatically finds individuals on Twitter by keywords expressing concepts, ideas and beliefs. Initial keywords are defined by users based on past or personal research. For example to identify the tribe of vegans one could start from keywords/hashtags such as "vegan", "plant based", "vegetarian", etc.. Subsequently the *Tribecreator* uses four search functions – analyzing Twitter profile descriptions, tweets, followers and friends networks – to identify key profiles that are typical for a certain tribe. These selected profiles are proposed to the user, who can manually select those to keep as what we call "tribal leaders".

The system visualizes the characteristics of such newly created tribes in two ways. First, *Tribecreator* draws a network of the members of the tribe, to provide a first representation of the most influential individuals. Second, it generates a hashtag word cloud, to see what the top tribe topics are and to suggest new keywords that could help refine the initial research. This initial representation can be adjusted recursively – refining the initial set of keywords at each step – to be more precise while selecting the key members (influential leaders) of each newly created tribe.

Analyzing the recurring concepts present in the tweets of influential leaders using deep learning with tensorflow (De Oliveira & Gloor, 2018), *Tribefinder* identifies the textual patterns that characterize each tribe and generates a specific tribal vocabulary. Subsequently, *Tribefinder* analyzes the language of the influential tribal leaders through deep learning to make the system learn how to associate random individuals with specific tribes (Gloor et al., 2018). Classifications are achieved by using word embeddings and LSTM (long short-term memory) models. Words are mapped into vectors, which are then used as input in the LSTM models (Greff, Srivastava, Koutnik, Steunebrink, & Schmidhuber, 2017). Once a tribe is predicted for each tweet, *Tribefinder* sums up the result to have a tribe allocation for the user.

The classification of short texts (as those found on Twitter) can be a challenge due to data sparseness (Vo & Ock, 2015), which makes the use of conventional machine learning and text mining algorithms difficult. Several approaches have been specifically designed for short text analysis, such as combining external text into target short text for classification (e.g., Banerjee, Ramanathan, & Gupta, 2007; Long, Chen, Zhu, & Zhang, 2012; Metzler, Dumais, & Meek, 2007; Sahami & Heilman, 2006) or using user-defined topics (e.g., Chen, Jin, & Shen, 2011; Phan, et al., 2011). In our work, we decided to adopt deep learning with word embeddings and found that length of tweets was not a problem, as far as they were combined in the analysis. What is more important is to have a sufficiently large number of tweets to combine – in particular while training the deep learning model, i.e. while defining a new tribe. For example, with our system is not possible to define a new tribe who has only three representatives that have few tweets. Conceptually *Tribefinder* is not dependent on deep learning, however we found better results applying deep learning compared to the more conventional information retrieval approaches referenced above.

**Figure 1.** *Tribefinder* system architecture

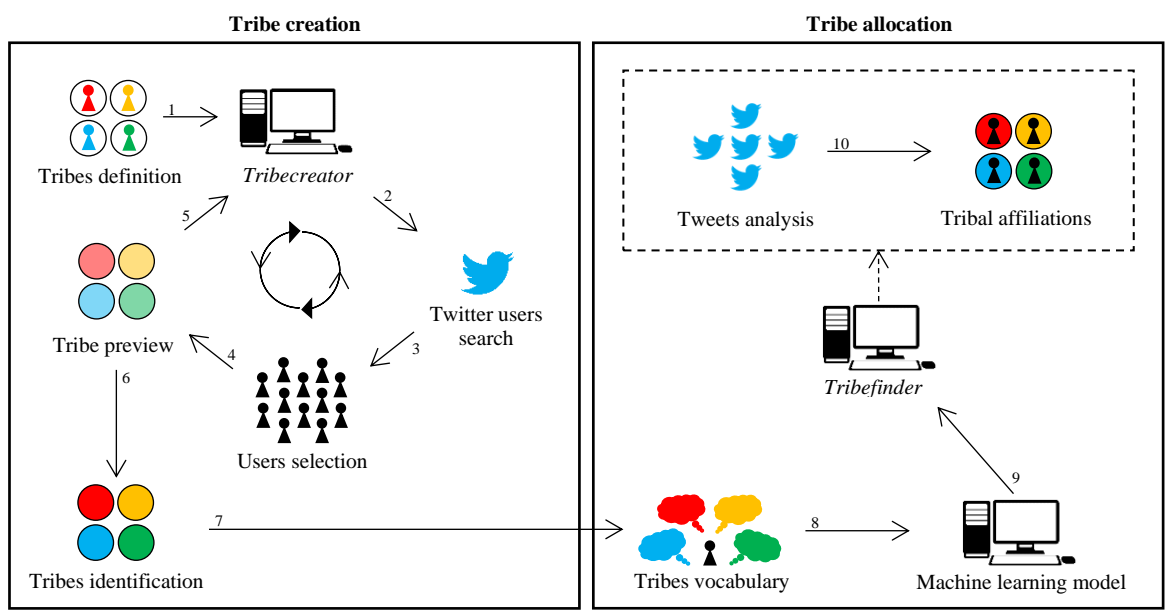

### 3.3. An example of tribe categories

In the experiment we present in Section 4, we used the *Tribefinder* to extract brand tribes and discuss their differences in terms of social dynamics and use of language, which may be useful practices for firms and their marketing campaigns. As mentioned above, we adopted the tribal categorizations presented in an earlier work that used the *Tribefinder* (Gloor, et al., 2019), without the ambition of theorizing or developing new ones. Accordingly, we have three macro-categories: *alternative realities*, *lifestyle*, and *recreation*.

For the category *alternative realities*, *Tribefinder* classifies each individual into one of four tribes: *fatherlanders*, *spiritualists*, *nerds*, and *treehuggers* (Gloor, et al., 2019). The *fatherlanders* are ultra-patriots who want to recreate the national states of the early twentieth century. The *spiritualists* mainly focus their attention on the spiritual side of things, the *nerds* are technocrats who believe in a global world ruled by capital and technology, and the *treehuggers* aim at protecting the environment.

In a second categorization, individuals are grouped into one of four different *lifestyles*: *fitness*, *sedentary*, *yolo*, and *vegan*. *Fitness* individuals are addicted to sport, while, on the contrary, those who are *sedentary* are characterized by very little or no physical exercise. *Yolo* individuals follow the motto "you only live once" and are inclined to impulsive and reckless behaviors. Lastly, the *vegans* are characterized by a plant-based diet, which avoids animal foods, as well as any animal product (Gloor, et al., 2019).

The third categorization, *recreation*, includes *fashion*, *art*, *travel*, and *sport*. The name of these categories clearly gives an indication of the preferences of individuals. *Fashion* individuals are addicted to popular or latest style of clothes, hair style, decoration and behavior, while *art* individuals are interested in any form of art. The category *travel* represents individuals who love travelling around the world, while the *sport* category includes those that love watching any type of sport (Gloor, et al., 2019).

## 4. Applications of the Tribefinder: Matching Tribes and Behaviors

*Tribefinder* is a powerful instrument that researchers and firms can use to identify, study, and exploit customers' tribal affiliations. The *Tribefinder* algorithm is more extensively described and validated in the work of Gloor, et al. (2019), who proved that this tool can achieve good classification accuracy (81.2% in the best case and 68.8% in the worst case). Among the different validation tests, they categorized the Twitter users who generated content related to several commercial brands. These brands were all quite popular and selected because they were mentioned in a sufficient number of tweets, and had the characteristic of attracting different brand-oriented tribes. This served to experimentally test the accuracy of the *Tribefinder* and verify whether its automatic classification method was aligned with expectations. For instance, the authors showed that a brand like Adidas was attracting more individuals from the *sport* tribe, than from the *art, travel* or *fashion* tribes.

In this study, we perform additional analyses which extend the use of the *Tribefinder* and combine it with the study of honest signals of communication (Gloor, 2017; Pentland, 2010). In particular, we use the tool to identify tribes and then explore the differences in the behavior of their members, in terms of social interaction patterns and use of language. This might be particularly useful for firms, to fully understand the characteristics and behaviors of their customers and then accordingly adjust their marketing campaigns.

### 4.1. A framework for the analysis of social dynamics

In order to explore the different behaviors and interaction patterns of the tribe members, we used metrics of social network and semantic analysis. In so doing, we adopted the framework proposed by previous literature (e.g., Antonacci, Fronzetti Colladon, Stefanini, & Gloor, 2017; Fronzetti Colladon & Gloor, 2018; Gloor, 2017; Gloor, Fronzetti Colladon, Giacomelli, Saran, & Grippa, 2017), which is based on the measurement of honest signals of communication (Pentland, 2010) and suggests describing the online social dynamics in terms of connectivity, interactivity, and language use. Honest signals are signals that "are either so costly to make or so difficult to suppress that they are reliable in signaling intentions" (Pentland, 2010, p. 2). To be considered as honest, signals must be processed unconsciously or be otherwise uncontrollable (Pentland, 2010). They are subtle patterns in human behaviors that reveal individual differences and intentions, without the biases of traditional surveys. These signals can be measured thanks to methods and tools of semantic and social network analysis, as subsequently outlined.

The degree of connectivity of a tribe is measured considering its social network and, specifically, network centrality (Freeman, 1978). We measured network centrality by means of two metrics, which are *degree* and *betweenness centrality*. The former consists of the number of direct links of a node. Applied to our specific case, degree centrality measures the number of individuals with whom a user interacts on Twitter. Betweenness centrality looks instead at the indirect links of a node, counting the number of times a user is in-between the paths that connect other users. This metric is thus computed by taking into account the shortest network paths that interconnect every possible pair of nodes and counts how many times these paths include a specific user (Wasserman & Faust, 1994).

The second dimension we considered to describe online social dynamics is interactivity, which provides an indication of the evolution of social dynamics over time. We used two metrics: *messages sent* and *rotating leadership*. The first metric is just the number of tweets of the user under investigation. To measure rotating leadership, we start with betweenness centrality described above. Specifically, rotating leadership consists of the number of oscillations in betweenness centrality of a given user – in other words, how many times betweenness centrality changes and reaches local maxima and minima in a given time interval. In practice, if the user is static, her/his rotating leadership is equal to 0. A user can be considered a rotating leader if s/he oscillates between central and peripheral positions, first activating conversations and then leaving them to other individuals in the network (Kidane & Gloor, 2007).

Finally, we used three indexes to analyze the use of language: *sentiment*, *emotionality*, and *complexity*. Sentiment is a measure of the positivity (or negativity) of tweets. We calculated this metric using a multi-lingual classifier based on a machine learning algorithm (see Brönnimann, 2013 for details; 2014) included in the software Condor (Gloor, 2017). Sentiment varies from 0 (i.e., a totally negative tweet) to 1 (i.e. totally positive tweet); values around 0.5 indicate a neutral tweet. Emotionality measures the deviation from neutral sentiment (Brönnimann, 2014). Tweets are more emotional when online users choose stronger words, and/or when individuals show opposing views, generating positive and negative posts. When emotionality is close to 0, users converge towards expressions that are neither positive or negative. The last metric is complexity, which provides an idea of how much the vocabulary used is complex, meaning that it deviates from common language. Complexity is measured as the probability of each word of a dictionary appearing in the tweets (Brönnimann, 2014). A word is considered as complex when it rarely appears in the context analyzed (and not in a general context); this means that words that are generally considered as rare might be instead common, in a specific discourse.

We calculated all these metrics with the software Condor (Gloor, 2017).

### 4.2. Experimental results

*Tribefinder* is a flexible tool, therefore it can be used to create different tribal categories, according to the context of analysis. However, in this study, we chose to be consistent with the original classifications presented in the work of Gloor, et al. (2019). To extend their previous research, we combined the tribe allocation results with metrics of social network analysis and text mining (e.g., Afful-Dadzie & Afful-Dadzie, 2017; Rekik, Kallel, Casillas, & Alimi, 2018), to illustrate a methodology useful to evaluate the distinguishing traits of each tribe. Such information can be valuable when formulating marketing strategies, or designing new products or communication campaigns. For instance, managers might be interested in increasing the sentiment of the tribes that are strategic to their business. Similarly, tribes with higher language complexity often introduce new ideas and scenarios, which might support the development of new products or the adjustment of existing ones.

For our experiment, we used a dataset made of about 192,800 tweets, posted by more than 66,000 users, which we collected by means of the social network and semantic analysis software Condor (Gloor, 2017). We collected tweets of April-May 2018 and we referred to the online discourse about 46 different brands (keeping the selection made by Gloor, et al., 2019). All the tweets were written in English.

The results of our analysis show that different tribes also have distinctive characteristics in terms of users' behaviors, interactions and language (see Figures 2-4 and Tables 1-3).

For the macro-category *alternative realities*, the one-way ANOVA (Table 1) reveals significant differences for all the dimensions considered (significance level of 99%), with the only exception of degree centrality (p-value = 0.837). As reported in Table 1 and Figure 2 , the most conservative – i.e., the *fatherlanders* – are the members with the lowest sentiment and the highest emotionality. They have fewer connections with the outside world and the tribe leaders are fairly static, with low values of rotating leadership. The language complexity of *fatherlanders* is low, indicating that they conform to the same concepts or set of words. Not surprisingly the *nerds* are those introducing more innovation in the discourse, with the highest value of language complexity. *Spiritualists* and *treehuggers*, on the other hand, show more commitment and optimistic views, having a higher sentiment, more messages sent, and leaders who rotate more.

**Figure 2.** Social and text metrics for the *alternative realities* macro-category (significant differences in means are marked with an asterisk)

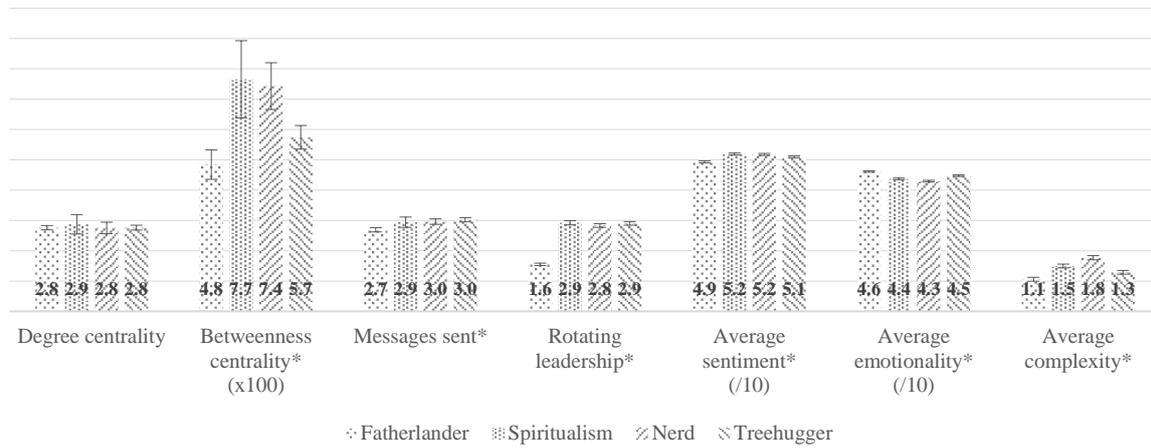

**Table 1.** Differences among *alternative realities* tribes

| | *Degree centrality* | | | | | | | | | |
|---|---|---|---|---|---|---|---|---|---|---|
| | One-way ANOVA | | | | | Post hoc analysis (Tukey HSD) | | | | |
| | Sum of squares | df | Mean square | F | Sig. | Mean | Fitness | Sedentary | Yolo | Vegan |
| Between groups | 58,911 | 3 | 19,637 | 0,284 | 0,837 | Fatherlander = 2,760 | | | | |
| Within groups | 1745365,441 | 25241 | 69,148 | | | Spiritualism = 2,870 | | | | |
| Total | 1745424,352 | 25244 | | | | Nerd = 2,750 | | | | |
| | | | | | | Treehugger = 2,760 | | | | |
| | *Betweenness centrality* | | | | | | | | | |
| | One-way ANOVA | | | | | Post hoc analysis (Tukey HSD) | | | | |
| | Sum of squares | df | Mean square | F | Sig. | Mean | Fitness | Sedentary | Yolo | Vegan |
| Between groups | 329374255,100 | 3 | 109791418,400 | 9,265 | 0,000 | Fatherlander = 483,947 | | *** | *** | |
| Within groups | 299105474400,000 | 25241 | 11849985,120 | | | Spiritualism = 765,240 | *** | | | *** |
| Total | 299434848600,000 | 25244 | | | | Nerd = 742,879 | *** | | | ** |
| | | | | | | Treehugger = 573,829 | | *** | ** | |
| | *Messages sent* | | | | | | | | | |
| | One-way ANOVA | | | | | Post hoc analysis (Tukey HSD) | | | | |
| | Sum of squares | df | Mean square | F | Sig. | Mean | Fitness | Sedentary | Yolo | Vegan |
| Between groups | 344,557 | 3 | 114,852 | 5,818 | 0,001 | Fatherlander = 2,690 | | ** | *** | *** |
| Within groups | 498262,430 | 25241 | 19,740 | | | Spiritualism = 2,940 | ** | | | |
| Total | 498606,987 | 25244 | | | | Nerd = 2,960 | *** | | | |
| | | | | | | Treehugger = 3,020 | *** | | | |
| | *Rotating leadership* | | | | | | | | | |
| | One-way ANOVA | | | | | Post hoc analysis (Tukey HSD) | | | | |
| | Sum of squares | df | Mean square | F | Sig. | Mean | Fitness | Sedentary | Yolo | Vegan |
| Between groups | 7342,151 | 3 | 2447,384 | 429,271 | 0,000 | Fatherlander = 1,550 | | *** | *** | *** |
| Within groups | 143905,426 | 25241 | 5,701 | | | Spiritualism = 2,920 | *** | | * | |
| Total | 151247,577 | 25244 | | | | Nerd = 2,830 | *** | * | | |
| | | | | | | Treehugger = 2,900 | *** | | | |
| | *Average sentiment* | | | | | | | | | |
| | One-way ANOVA | | | | | Post hoc analysis (Tukey HSD) | | | | |
| | Sum of squares | df | Mean square | F | Sig. | Mean | Fitness | Sedentary | Yolo | Vegan |
| Between groups | 2,433 | 3 | 0,811 | 46,066 | 0,000 | Fatherlander = 0,493 | | *** | *** | *** |
| Within groups | 444,359 | 25241 | 0,018 | | | Spiritualism = 0,519 | *** | | | *** |
| Total | 446,791 | 25244 | | | | Nerd = 0,517 | *** | | | *** |
| | | | | | | Treehugger = 0,510 | *** | *** | *** | |
| | *Average emotionality* | | | | | | | | | |

|  | One-way ANOVA | | | | | Post hoc analysis (Tukey HSD) | | | | |
| --- | --- | --- | --- | --- | --- | --- | --- | --- | --- | --- |
|  | Sum of squares | df | Mean square | F | Sig. | Mean | Fitness | Sedentary | Yolo | Vegan |
| Between groups | 3,397 | 3 | 1,132 | 94,158 | 0,000 | Fatherlander = 0,461 |  | *** | *** | *** |
| Within groups | 303,549 | 25241 | 0,012 |  |  | Spiritualism = 0,437 | *** |  | *** | *** |
| Total | 306,946 | 25244 |  |  |  | Nerd = 0,429 | *** | *** |  | *** |
|  |  |  |  |  |  | Treehugger = 0,447 | *** | *** | *** |  |

*Average complexity*

|  | One-way ANOVA | | | | | Post hoc analysis (Tukey HSD) | | | | |
| --- | --- | --- | --- | --- | --- | --- | --- | --- | --- | --- |
|  | Sum of squares | df | Mean square | F | Sig. | Mean | Fitness | Sedentary | Yolo | Vegan |
| Between groups | 1672,732 | 3 | 557,577 | 80,711 | 0,000 | Fatherlander = 1,057 |  | *** | *** | *** |
| Within groups | 174373,346 | 25241 | 6,908 |  |  | Spiritualism = 1,495 | *** |  | *** | *** |
| Total | 176046,078 | 25244 |  |  |  | Nerd = 1,769 | *** | *** |  | *** |
|  |  |  |  |  |  | Treehugger = 1,287 | *** | *** | *** |  |

For the macro-category *lifestyle*, significant differences at the 0.01 level emerge for betweenness centrality, rotating leadership, average sentiment, average emotionality and average complexity (see Table 2). Conversely, the one-way ANOVA reveals that *lifestyle* tribes do not differ in terms of degree centrality (p-value = 0.116) and messages sent (p-value = 0.327). Looking at Table 2 and Figure 3, we find that members of the *fitness* tribe have the highest sentiment and more brokerage power, being often positioned in-between the paths that interconnect the other social actors. They stimulate the discourse, but their positions are rather static, having low values of rotating leadership. They are also more emotionally stable, with less alternation of positive and negative sentiment. By contrast, the *vegan* tribe is the most dynamic, with high values of rotating leadership, and lower sentiment. The *fitness* and *sedentary* tribes have more connections that span across different social groups.

**Figure 3.** Social and text metrics for the *lifestyle* macro-category (significant differences in means are marked with an asterisk)

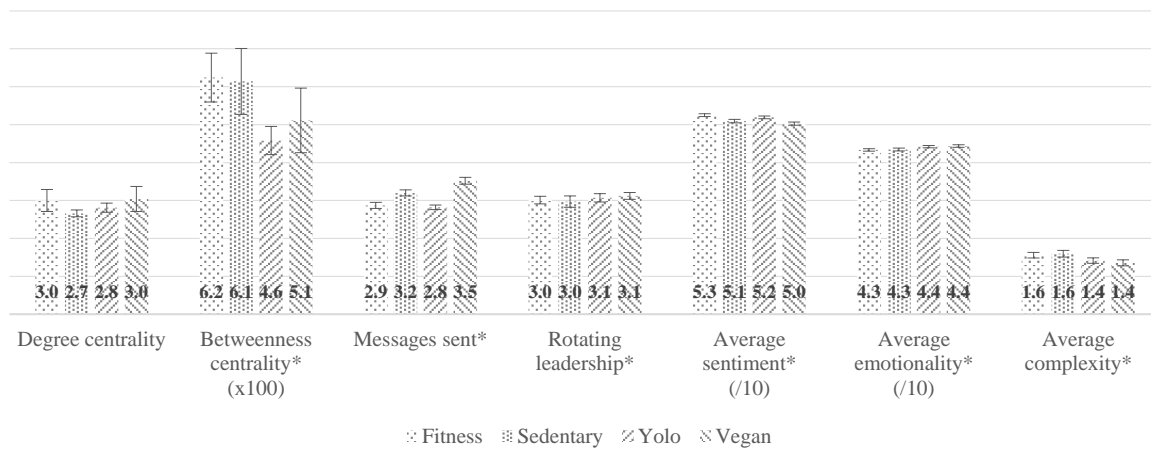

**Table 2.** Differences among *lifestyle* tribes

*Degree centrality*

| | One-way ANOVA | | | | | Post hoc analysis (Tukey HSD) | | | | |
|---|---|---|---|---|---|---|---|---|---|---|
| | Sum of squares | df | Mean square | F | Sig. | Mean | Fitness | Sedentary | Yolo | Vegan |
| Between groups | 383,296 | 3 | 127,765 | 1,971 | 0,116 | Fitness = 3,000 | | | | |
| Within groups | 1176692,736 | 18153 | 64,821 | | | Sedentary = 2,660 | | | | |
| Total | 1177076,032 | 18156 | | | | Yolo = 2,810 | | | | |
| | | | | | | Vegan = 3,040 | | | | |

*Betweenness centrality*

| | One-way ANOVA | | | | | Post hoc analysis (Tukey HSD) | | | | |
|---|---|---|---|---|---|---|---|---|---|---|
| | Sum of squares | df | Mean square | F | Sig. | Mean | Fitness | Sedentary | Yolo | Vegan |
| Between groups | 93866737,130 | 3 | 31288912,380 | 5,860 | 0,001 | Fitness = 624,215 | | | *** | * |
| Within groups | 96925751620,000 | 18153 | 5339379,255 | | | Sedentary = 614,084 | | | *** | |
| Total | 97019618360,000 | 18156 | | | | Yolo = 458,145 | *** | *** | | |
| | | | | | | Vegan = 511,276 | * | | | |

*Messages sent*

| | One-way ANOVA | | | | | Post hoc analysis (Tukey HSD) | | | | |
|---|---|---|---|---|---|---|---|---|---|---|
| | Sum of squares | df | Mean square | F | Sig. | Mean | Fitness | Sedentary | Yolo | Vegan |
| Between groups | 52,953 | 3 | 17,651 | 1,152 | 0,327 | Fitness = 3,010 | | | | |
| Within groups | 278243,022 | 18153 | 15,328 | | | Sedentary = 2,970 | | | | |
| Total | 278295,975 | 18156 | | | | Yolo = 3,070 | | | | |
| | | | | | | Vegan = 3,120 | | | | |

*Rotating leadership*

| | One-way ANOVA | | | | | Post hoc analysis (Tukey HSD) | | | | |
|---|---|---|---|---|---|---|---|---|---|---|
| | Sum of squares | df | Mean square | F | Sig. | Mean | Fitness | Sedentary | Yolo | Vegan |
| Between groups | 1.431,323 | 3 | 477,108 | 64,769 | 0,000 | Fitness = 2,870 | | *** | | *** |
| Within groups | 133720,655 | 18153 | 7,366 | | | Sedentary = 3,200 | *** | | *** | *** |
| Total | 135151,979 | 18156 | | | | Yolo = 2,820 | | *** | | *** |
| | | | | | | Vegan = 3,520 | *** | *** | *** | |

*Average sentiment*

| | One-way ANOVA | | | | | Post hoc analysis (Tukey HSD) | | | | |
|---|---|---|---|---|---|---|---|---|---|---|
| | Sum of squares | df | Mean square | F | Sig. | Mean | Fitness | Sedentary | Yolo | Vegan |
| Between groups | 1,355 | 3 | 0,452 | 24,431 | 0,000 | Fitness = 0,525 | | *** | | *** |
| Within groups | 335,668 | 18153 | 0,018 | | | Sedentary = 0,509 | *** | | *** | |
| Total | 337,024 | 18156 | | | | Yolo = 0,519 | | *** | | *** |
| | | | | | | Vegan = 0,503 | *** | | *** | |

*Average emotionality*

|  | One-way ANOVA | | | | | Post hoc analysis (Tukey HSD) | | | | |
|---|---|---|---|---|---|---|---|---|---|---|
|  | Sum of squares | df | Mean square | F | Sig. | Mean | Fitness | Sedentary | Yolo | Vegan |
| Between groups | 0,368 | 3 | 0,123 | 9,349 | 0,000 | Fitness = 0,433 |  |  | *** | *** |
| Within groups | 238,460 | 18153 | 0,013 |  |  | Sedentary = 0,434 |  |  | ** | *** |
| Total | 238,828 | 18156 |  |  |  | Yolo = 0,442 | *** | ** |  |  |
|  |  |  |  |  |  | Vegan = 0,444 | *** | *** |  |  |

*Average complexity*

|  | One-way ANOVA | | | | | Post hoc analysis (Tukey HSD) | | | | |
|---|---|---|---|---|---|---|---|---|---|---|
|  | Sum of squares | df | Mean square | F | Sig. | Mean | Fitness | Sedentary | Yolo | Vegan |
| Between groups | 165,748 | 3 | 55,249 | 7,766 | 0,000 | Fitness = 1,557 |  |  | ** | *** |
| Within groups | 129137,279 | 18153 | 7,114 |  |  | Sedentary = 1,595 |  |  | *** | *** |
| Total | 129303,027 | 18156 |  |  |  | Yolo = 1,414 | ** | *** |  |  |
|  |  |  |  |  |  | Vegan = 1,355 | *** | *** |  |  |

Lastly, for the macro-category *recreation*, we find highly significant differences among tribes on messages sent, rotating leadership, average sentiment, average emotionality and average complexity (Table 3). Marginally significant differences emerge for degree centrality (p-value = 0.097), while tribes do not differ in terms of betweenness centrality (p-value = 0.624). We see more dynamism in the *travel* tribe (leaders rotate more), even if the highest volume of tweets is produced by the *sport* tribe. Members of the *sport* tribe have high brokerage power, tweet a lot, and are more connected. Nevertheless, the leaders of this tribe are fairly static, they keep their dominant position, centralizing the conversations. Sentiment is higher in the *fashion* tribe, where the discourse is also more emotional. Tweeting about travel destinations introduces more variety in the online interaction, thus leading to a higher value of complexity.

**Figure 4.** Social and text metrics for the *recreation* macro-category (significant differences in means are marked with an asterisk)

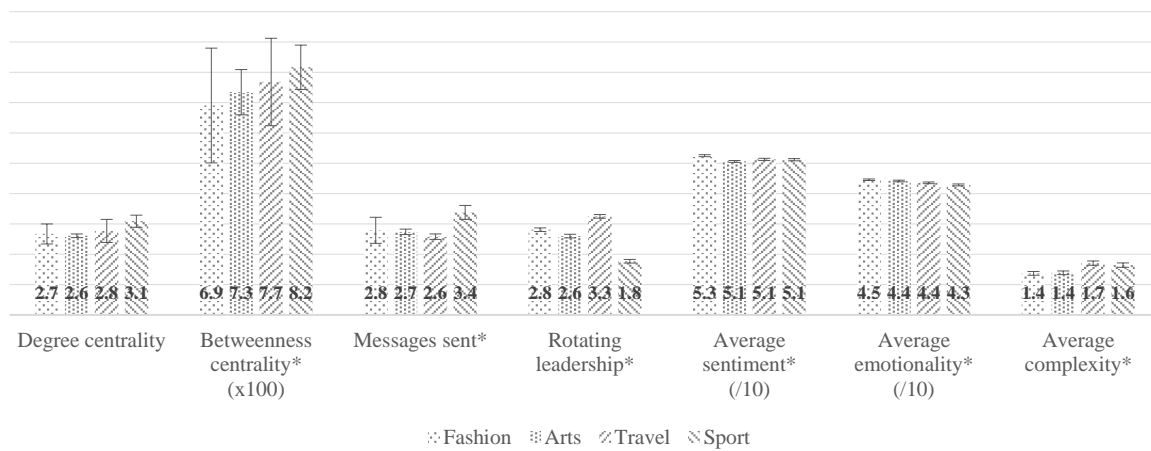

**Table 3.** Differences among *recreation* tribes

| | *Degree centrality* | | | | | | | | | |
|---|---|---|---|---|---|---|---|---|---|---|
| | One-way ANOVA | | | | | | Post hoc analysis (Tukey HSD) | | | |
| | Sum of squares | df | Mean square | F | Sig. | Mean | Fitness | Sedentary | Yolo | Vegan |
| Between groups | 823,503 | 3 | 274,501 | 2,108 | 0,097 | Fashion = 2,670 | | | | |
| Within groups | 3396310,271 | 26076 | 130,247 | | | Arts = 2,610 | | | | * |
| Total | 3397133,774 | 26079 | | | | Travel = 2,770 | | | | |
| | | | | | | Sport = 3,090 | | * | | |
| | *Betweenness centrality* | | | | | | | | | |
| | One-way ANOVA | | | | | | Post hoc analysis (Tukey HSD) | | | |
| | Sum of squares | df | Mean square | F | Sig. | Mean | Fitness | Sedentary | Yolo | Vegan |
| Between groups | 52831105,200 | 3 | 17610368,400 | 0,586 | 0,624 | Fashion = 690,594 | | | | |
| Within groups | 783495332200,000 | 26076 | 30046607,310 | | | Arts = 734,467 | | | | |
| Total | 783548163300,000 | 26079 | | | | Travel = 768,263 | | | | |
| | | | | | | Sport = 816,644 | | | | |
| | *Messages sent* | | | | | | | | | |
| | One-way ANOVA | | | | | | Post hoc analysis (Tukey HSD) | | | |
| | Sum of squares | df | Mean square | F | Sig. | Mean | Fitness | Sedentary | Yolo | Vegan |
| Between groups | 2174,262 | 3 | 724,754 | 6,640 | 0,000 | Fashion = 2,790 | | | | *** |
| Within groups | 2846097,137 | 26076 | 109,146 | | | Arts = 2,740 | | | | *** |
| Total | 2848271,399 | 26079 | | | | Travel = 2,580 | | | | *** |
| | | | | | | Sport = 3,380 | *** | *** | *** | |
| | *Rotating leadership* | | | | | | | | | |
| | One-way ANOVA | | | | | | Post hoc analysis (Tukey HSD) | | | |
| | Sum of squares | df | Mean square | F | Sig. | Mean | Fitness | Sedentary | Yolo | Vegan |
| Between groups | 6866,004 | 3 | 2288,668 | 363,352 | 0,000 | Fashion = 2,810 | | *** | *** | *** |
| Within groups | 164246,469 | 26076 | 6,299 | | | Arts = 2,600 | *** | | *** | *** |
| Total | 171112,472 | 26079 | | | | Travel = 3,250 | *** | *** | | *** |
| | | | | | | Sport = 1,770 | *** | *** | *** | |
| | *Average sentiment* | | | | | | | | | |
| | One-way ANOVA | | | | | | Post hoc analysis (Tukey HSD) | | | |
| | Sum of squares | df | Mean square | F | Sig. | Mean | Fitness | Sedentary | Yolo | Vegan |
| Between groups | 1,322 | 3 | 0,441 | 22,799 | 0,000 | Fashion = 0,525 | | *** | *** | *** |
| Within groups | 504,159 | 26076 | 0,019 | | | Arts = 0,506 | *** | | ** | * |
| Total | 505,481 | 26079 | | | | Travel = 0,513 | *** | ** | | |
| | | | | | | Sport = 0,512 | *** | * | | |
| | *Average emotionality* | | | | | | | | | |

|  | One-way ANOVA | | | | | Post hoc analysis (Tukey HSD) | | | | |
|--|---|---|---|---|---|---|---|---|---|---|
|  | Sum of squares | df | Mean square | F | Sig. | Mean | Fitness | Sedentary | Yolo | Vegan |
| Between groups | 1,021 | 3 | 0,340 | 26,299 | 0,000 | Fashion = 0,446 |  |  | *** | *** |
| Within groups | 337,587 | 26076 | 0,013 |  |  | Arts = 0,442 |  |  | ** | *** |
| Total | 338,609 | 26079 |  |  |  | Travel = 0,436 | *** | ** |  | *** |
|  |  |  |  |  |  | Sport = 0,429 | *** | *** | *** |  |

*Average complexity*

|  | One-way ANOVA | | | | | Post hoc analysis (Tukey HSD) | | | | |
|--|---|---|---|---|---|---|---|---|---|---|
|  | Sum of squares | df | Mean square | F | Sig. | Mean | Fitness | Sedentary | Yolo | Vegan |
| Between groups | 593,072 | 3 | 197,691 | 26,541 | 0,000 | Fashion = 1,369 |  |  | *** | *** |
| Within groups | 194227,894 | 26076 | 7,449 |  |  | Arts = 1,389 |  |  | *** | *** |
| Total | 194820,967 | 26079 |  |  |  | Travel = 1,708 | *** | *** |  |  |
|  |  |  |  |  |  | Sport = 1,642 | *** | *** |  |  |

The analysis we presented concerns the discourse about 46 well-known brands, whose list was taken from Gloor, et al. (2019). This sample is therefore representative of a heterogeneous spectrum of consumers and stakeholders, without being focused on a specific business sector. It served the purpose of illustrating a methodology useful to monitor tribe behaviors, before planning appropriate interventions. However, it might be interesting for brand managers, to replicate our methodology only collecting tweets about their brand and possibly about its competitors. Accordingly, we present an additional example of an analysis we made by considering the *fashion* tribe in the Twitter discourse of the following four brands: Gucci, Chanel, Dior, Louis Vuitton. It is not surprising that the fashion tribe is predominant for these brands.

**Figure 5.** Differences in fashion brands tribes (significant differences in means are marked with an asterisk)

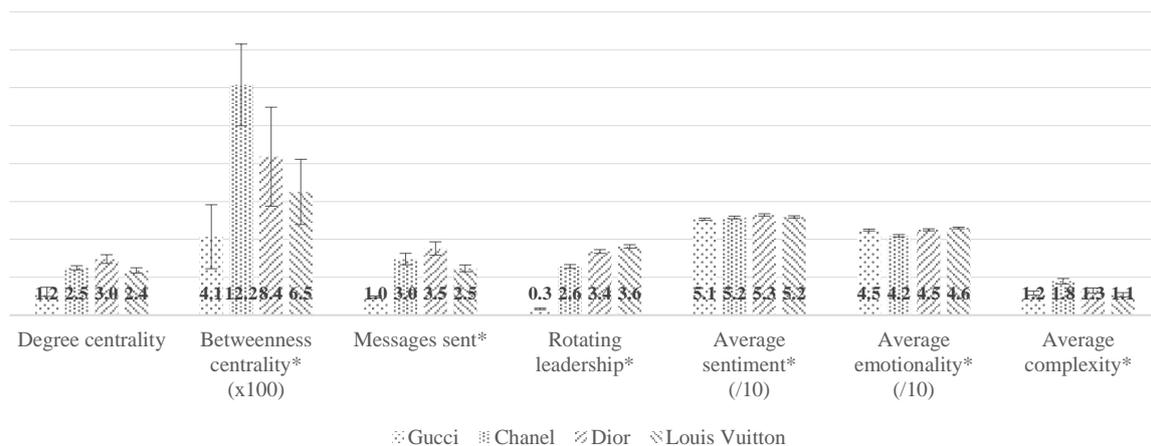

Figure 5 shows that even if the tribe is kept the same for the different tribes, behaviors significantly differ when the discourse regards one brand or another. In this example, the brand Gucci receives less interest and appreciation: the sentiment is lower than for the other brands, the discourse is rather flat (low complexity), and the volume of tweets is also small. Chanel and Dior, on the other hand, have a more heterogeneous discourse, their tribes are more connected, and the volume of tweets is significantly higher. Dior has also the highest sentiment.

It is important to notice that the approach we adopted here strongly differs from an analysis of tweets carried out without the identification of tribes; a limitation that characterizes many studies (e.g., La Bella, Fronzetti Colladon, Battistoni, Castellan, & Francucci, 2018). Generally speaking, tweets about a company may be about very different topics – such as its financial performance, its products, or the latest news about the board members. Using *Tribefinder* permits users to identify more homogeneous groups of individuals (tribes) who discuss specific topics and show similar interests. For instance, an analyst could find a neutral average sentiment when looking at the discourse about a firm; however, this information might be misleading as the sentiment could be positive for the group of individuals who talk about the firm products, and negative in the tribe of financial analysts who make predictions about the company's future performance.

## 5. Conclusion

This paper presents *Tribefinder*, a new instrument to identify (virtual) tribes of Twitter users. By analyzing Twitter users' tweets and vocabulary through a combination of word embeddings and LSTM models, *Tribefinder* is able to understand to what tribes a sample of individuals active on Twitter belong. In our experiment, we considered three tribal macro-categories: alternative realities, lifestyle, and recreation. These categorizations are taken as examples to demonstrate the instrument and illustrate its functionality. Tribefinder can be easily extended according to the interests of its users. Using the methodology described in this paper, *Tribefinder* can be extended to identify other user-defined tribal macro-categories. Moreover, although our analysis was limited to Twitter, *Tribefinder* can be applied to alternative social media platforms, for instance Reddit, as we plan to do in future work.

Apart from illustrating this new system, we make a step forward from the previous work of Gloor, et al. (2019), by illustrating a methodology apt at evaluating the distinguishing traits of the tribes identified through this new instruments. Specifically, we described tribes' social dynamics in terms of connectivity, interactivity, and language use, by means of metrics of social network and semantic analysis.

### 5.1. Theoretical contributions

We are convinced that the instrument we propose may have important implications to advance the academic conversation. First, our work contributes to advancing past and current state of the art by offering a new instrument that is able to identify individuals' tribal affiliations and extract richer data on virtual tribes from the analysis of digital communication than existing methodologies do. The *Tribefinder* capabilities set the ground for further advancements in the field of marketing research. Our work allows researchers to overcome the limitations of the traditional approaches that have been used so far to identify and study consumer tribes (e.g., focus groups, interviews, or surveys), which may be time consuming and do not allow an automatic identification of virtual tribes and their characteristics. In addition, combining *Tribefinder* with the measurement of honest signals of communication (Gloor, 2017; Pentland, 2010), scholars have at their disposal new information that may be used to advance research on tribal marketing. Indeed, *Tribefinder* and the measurement of honest signal of communication provide useful information that may be exploited to understand how marketing actions can be used in a strategic way, an aspect that has not been explored much despite its renewed importance (Cova & Cova, 2002). In other words, this methodology offers a foundation for future research on *social media marketing* (see Alalwan, et al., 2017; Dwivedi, et al., 2015 for a comprehensive literature reviewes); it allows for a potential extension of traditional marketing solutions (Addis & Podesta, 2005; Canniford, 2011) and helps firms understand the essence of their consumers' tribes (Moutinho, et al., 2007). Indeed, previous research mainly adopted questionnaires, qualitative interviews or basic content analyses to materials posted on social media (Alalwan, et al., 2017). Recently several scholars made attempts in adopting new methodologies to analyze tweets and derive information from them (e.g., Aswani, Kar, Ilavarasan, & Dwivedi, 2018; Fronzetti Colladon & Gloor, 2018; Grover, et al., 2018; Zhang, Fuehres, & Gloor, 2011). However, to the best of our knowledge, little effort has been invested into the automatic identification of Twitter users' tribal affiliations. In addition, limited work exists that examines the organizational use of Twitter (Burton & Soboleva, 2011; Martínez-Rojas, et al., 2018). We thus make a step further in this direction by providing an easy-to-use instrument designed to classify users into tribes by analyzing their tweets, and thus offering firms a foundation to leverage Twitter for their organizational goals. Finally, our work contributes to general research on social media mining (e.g., Grover, et al., 2018; Jeong, et al., 2017; Singh, et al., 2017a) by presenting a new instrument to analyze, categorize and extract information from user-generated content.

### 5.2. Managerial implications

Apart from its academic implications, we are confident that *Tribefinder* will be extremely useful from a practitioner's point of view. Indeed, we know from literature that *tribes* are of fundamental importance for firms' survival (e.g., Holzweber, et al., 2015), and especially for marketing (e.g., Goulding, et al., 2013; Kozinets, 1999). Tribes' characteristics affect the success of a marketing campaign (Cova & Cova, 2002). Moreover, Internet and the diffusion of social media platforms contributed to changing marketing paradigms (Burton & Soboleva, 2011). This demands new marketing strategies (Addis & Podesta, 2005; Canniford, 2011) based on the analysis of the tribes interested in the products and services a firm offers (Cova & Cova, 2002; Moutinho, et al., 2007). Starting from this premise, firms and their marketers can directly use *Tribefinder* to get a better knowledge of their customers who interact on Twitter. Firstly, firms can identify those Twitter users who form their brand community, studying their characteristics and behaviors. Secondly, analyzing the identified virtual tribes, firms can more easily measure the efficiency of their marketing campaigns, and adjust their marketing strategies (Cova, 1996; Holzweber, et al., 2015; Moutinho, et al., 2007). In this regard, our approach to measure honest signal of communication may be of key importance. By analyzing connectivity, interactivity, and language use of tribe members, managers can understand the tribe's traits for designing new products or communication campaigns. In addition, firms may collect similar information about their competitors. Ultimately, the use of *Tribefinder* can contribute to improving companies' competitive advantage and performance.

### 5.3. Limitations and future research direction

Our paper is not devoid of limitations, which open areas for future research. In presenting and testing our system, we used only Twitter. While we already mentioned that this is not intended to be a limitation, as *Tribefinder* can be adapted to other social media (e.g., Reddit or email communication), a formal test of these new uses is advised. This would allow researchers to empirically prove its generalizability. In a similar vein, presenting its first application we focused on three specific tribal macro-categories – i.e., alternative realities, lifestyle, and recreation. Although we see value in these categories – as they allow to capture important characteristics of individuals who interact on social media – we invite scholars to extend our work by defining and testing other macro-categories.

In analyzing Twitter short texts we decided to use deep learning with word embeddings. However, alternative methodologies exists, which could be exploited in future research to assess the quality of our choice and potentially improve our approach. Similarly, further research could refer to other metrics of honest signals, such as the measurement of the *average response time* taken by a user to respond to a tweet where s/he is mentioned. The quality of our data did not allow the computation of this metric of interactivity because timestamps were sometimes not properly recorded and many tweets were left without answers or mentions by other users.

Finally, once *Tribefinder* reaches a sufficient diffusion, its impact on business practices and performance could be investigated. Apart from proving its importance for firms and practitioners in general, it could stimulate new research on tribal marketing that exploits rich data collected through our new system.